\documentclass[10pt,twocolumn,letterpaper]{article}

\usepackage[pagenumbers]{iccv} % To force page numbers, e.g. for an arXiv version

\usepackage{mdwmath}
\usepackage{eqparbox}
\usepackage{epsfig}
\usepackage{graphicx}
\usepackage{amsmath}
\usepackage{amssymb}
\usepackage{pifont}
\usepackage{url}            % simple URL typesetting
\usepackage{booktabs}       % professional-quality tables
\usepackage{tabu}           % tabular
\usepackage{multirow}
\usepackage{graphicx}
\usepackage{algorithm}
\usepackage{algorithmicx}
\usepackage{algpseudocode}
\usepackage{amsmath,amssymb}
\usepackage{array}

\newcommand{\cmark}{\ding{51}}%
\newcommand{\xmark}{\ding{55}}%
\usepackage{cite}

\usepackage{makecell}
\usepackage{tabularx}
\usepackage{pifont}
\usepackage{multicol}
\usepackage{array}
\usepackage{adjustbox}

\usepackage{url}            % simple URL typesetting
\usepackage{booktabs}       % professional-quality tables
\usepackage{amsfonts}       % blackboard math symbols
\usepackage{nicefrac}       % compact symbols for 1/2, etc.
\usepackage{microtype}      % microtypography
\usepackage{graphicx}
\usepackage{amsmath}
\usepackage{bm}
\usepackage{multirow}
\usepackage{tabu}
\usepackage{siunitx}
\usepackage{adjustbox}
\usepackage{subcaption}
\usepackage{siunitx}
\usepackage{colortbl}
\usepackage{color}
\usepackage{pifont}
\definecolor{Gray}{gray}{0.9}
\definecolor{Lightorange}{RGB}{255,214,169}
\definecolor{Cyan}{rgb}{0.88,1,1}
\definecolor{reminder}{RGB}{255,0,0}

\usepackage{tabu}           % tabular
\usepackage{multirow}
\usepackage{booktabs}
\usepackage{makecell}
\usepackage{pifont}

\newcommand{\paragrapha}[2][3pt]{\vspace{#1}\noindent\textbf{#2}}

\newcolumntype{x}[1]{>{\centering\arraybackslashå}p{#1pt}}
\newlength\savewidth

\newcommand{\PreserveBackslash}[1]{\let\temp=\\#1\let\\=\temp}
\newcolumntype{C}[1]{>{\PreserveBackslash\centering}p{#1}}
\newcolumntype{L}[1]{>{\PreserveBackslash\raggedright}p{#1}}

\usepackage{url}

\definecolor{iccvblue}{rgb}{0.21,0.49,0.74}
\usepackage[pagebackref,breaklinks,colorlinks,allcolors=iccvblue]{hyperref}

\title{\emph{\color{Orange}Ola}: Pushing the Frontiers of Omni-Modal Language Model}

\def\spaces{~~~~~~}
\author{Zuyan Liu\textsuperscript{1,2}\thanks{Authors contributed equally to this research.~~\textsuperscript{\dag}Corresponding authors.}\spaces{}Yuhao Dong\textsuperscript{3,2}\footnotemark[1]\spaces{}Jiahui Wang\textsuperscript{1}\spaces{} \\
Ziwei Liu\textsuperscript{3}\spaces{}Winston Hu\textsuperscript{2}\spaces{}Jiwen Lu\textsuperscript{1}$^{\dagger}$\spaces{}Yongming Rao\textsuperscript{2,1}$^{\dagger}$ \\ \\
\textsuperscript{1}~Tsinghua University\spaces{}\textsuperscript{2}~Tencent Hunyuan Research\spaces{}\textsuperscript{3}~S-Lab, NTU\\ \\
\textbf{\color{Orange}\url{https://ola-omni.github.io/}}
}

\begin{document}
% \maketitle

\twocolumn[{
    \renewcommand\twocolumn[1][]{#1}
    \maketitle
    \begin{center}
  % \fbox{\rule{0pt}{1.6in} \rule{0.9\linewidth}{0pt}}
        \includegraphics[width=1.0\linewidth]{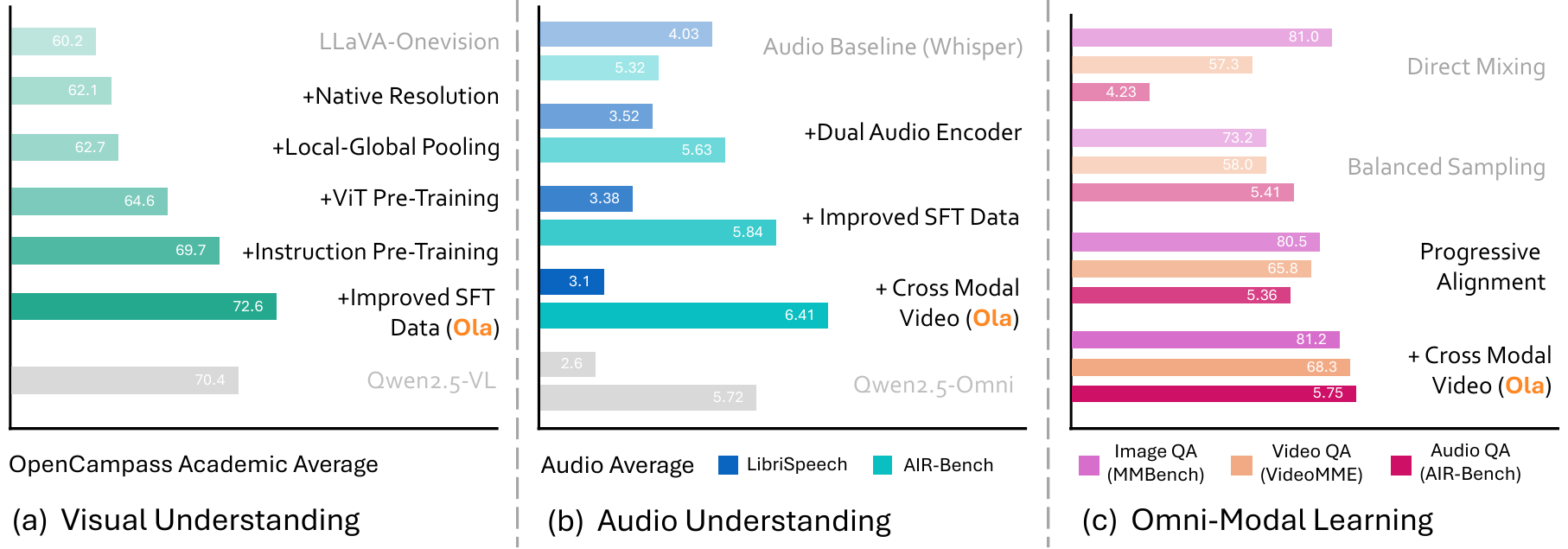}
        \captionof{figure}{ \textbf{Roads to \textit{Ola} Omni-Modal Language Model.} We illustrate our innovations on visual and audio understanding in subfigures (a) and (b). Benefiting from the improved architectures and data, \textit{Ola} achieves comparable performance with specialized models, outperforming state-of-the-art omni-modal models. We carefully design the training strategy for omni-modal models based on cross-modality and progressive alignment, as illustrated in subfigure (c). We compare our approach with two common baselines when training omni-modal models: 1) direct mixing where all instruction tuning data is merged and trained in a single stage, and 2) balanced sampling where we upsample certain sources to make the training data more balanced among modalities. }
        \label{fig:compare}
    \end{center}
}]

\let\thefootnote\relax\footnotetext{*~Equal Contribution.~~\textsuperscript{\dag}~Corresponding authors.}

% \begin{figure*}[!h]
%   \centering
%   \fbox{\rule{0pt}{1.6in} \rule{0.9\linewidth}{0pt}}
%    %\includegraphics[width=0.8\linewidth]{egfigure.eps}
%    \caption{Example of caption.
%    It is set in Roman so that mathematics (always set in Roman: $B \sin A = A \sin B$) may be included without an ugly clash.}
%    \label{fig:overview}
% \end{figure*}

\begin{abstract}

\begin{figure*}
% \fbox{\rule{0pt}{1.6in} \rule{0.9\linewidth}{0pt}}
    \includegraphics[width=1.0\linewidth]{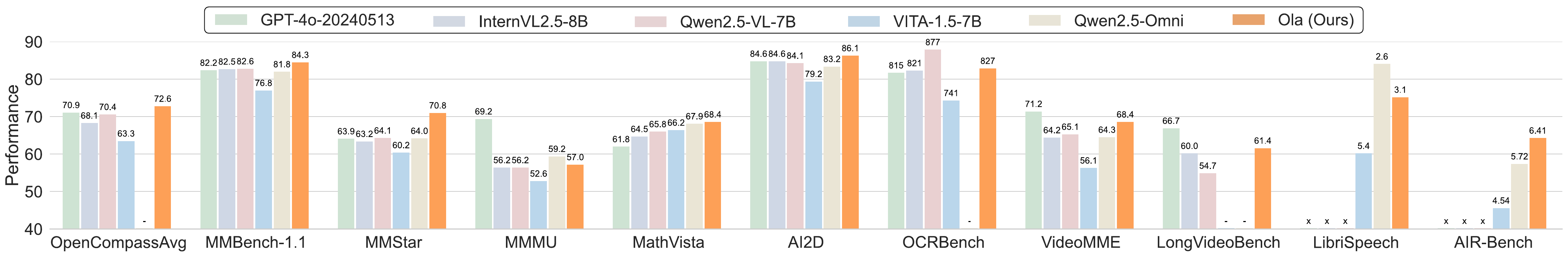}
    \captionof{figure}{\small \textbf{\textit{Ola} Pushes the Frontiers of the Omni-Modal Language Model across Image, Video and Audio Understanding Benchmarks. } We compare \emph{Ola} with existing state-of-the-art open-sourced multimodal models (InternVL-2.5~\citep{chen2024intern25} and Qwen2.5-VL~\citep{qwen2.5}), omni-modal models (Qwen2.5-Omni~\citep{xu2025qwen2} and VITA-1.5~\citep{fu2025vita1_5}), and GPT-4o~\citep{GPT4o}. For fair comparisons, we select around 7B versions of existing MLLMs. \emph{Ola} can achieve outperforming performance against omni-modal and specialized MLLMs in all modalities. ``$\times$'' indicates that the model is not capable of the task and ``$-$'' indicates the result is lacking. The score for LibriSpeech is inverted as lower is better for the WER metric.}
    \label{fig:teaser}
\end{figure*}

Recent advances in large language models, particularly following GPT-4o, have sparked increasing interest in developing omni-modal models capable of understanding more modalities. While some open-source alternatives have emerged, there is still a notable lag behind specialized single-modality models in performance. 
In this paper, we present \textbf{\textit{Ola}}, an \textbf{\textit{O}}mni-modal \textbf{\textit{La}}nguage model that achieves competitive performance across image, video, and audio understanding compared to specialized counterparts, pushing the frontiers of the omni-modal language model to a large extent. 
We conduct a comprehensive exploration of architectural design, data curation, and training strategies essential for building a robust omni-modal model. \textit{Ola} incorporates advanced visual understanding and audio recognition capabilities through several critical and effective improvements over mainstream baselines. Moreover, we rethink inter-modal relationships during omni-modal training, emphasizing cross-modal alignment with video as a central bridge, and propose a progressive training pipeline that begins with the most distinct modalities and gradually moves towards closer modality alignment.
Extensive experiments demonstrate that \textit{Ola} surpasses existing open omni-modal LLMs across all modalities while achieving highly competitive performance compared to state-of-the-art specialized models of similar sizes. We aim to make \textit{Ola} a fully open omni-modal understanding solution to advance future research in this emerging field. Model weights, code, and data are open-sourced at \url{https://github.com/Ola-Omni/Ola}. 

\end{abstract}  
\section{Introduction}
\label{sec:intro}

% \begin{figure*}[t]
% \centering
% \includegraphics[width=\textwidth]{figs/fig0-new.pdf} \vspace{-10pt}
% \caption{\textbf{\textit{Ola} Architecture. }\textit{Ola} supports omni-modal inputs including text, image, video, and audio, capable of processing the inputs simultaneously with competitive performance on understanding tasks for all these modalities. Meanwhile, \textit{Ola} supports user-friendly real-time streaming decoding for texts and speeches thanks to the text detokenizer and the speech decoder. }
% \label{fig:sota}\vspace{-10pt}
% \end{figure*}
Multi-Modal Large Language Models are drawing increasing attention owing to their strong instruction-following capabilities and abundant knowledge of handling complex inputs including texts, images, videos, and audio. Based on the strong performance of open-sourced large language models~\citep{qwen2,young2024yi}, extensive research has been done on connecting specific modalities with language responses~\citep{tong2024cambrian,li2024llavaov,chen2024internvl,qwen2vl,lin2023videollava,qian2024streaming}. Recently, the success of GPT-4o~\citep{GPT4o} and Gemini~\citep{reid2024gemini} aiming at supporting more modalities in Large Language Models inspires researchers to take one important steps towards omni models that understand all the inputs in one model. 

The core challenges in training omni-modal Large Language Models lie in the modeling of modalities in various distributions, and the design of an effective learning pipeline to achieve competitive, balanced performance on all the supported tasks. Several attempts have been made to overcome the difficulty of omni-modal models~\citep{fu2024vita,fang2024llamaomni,xie2024miniomni2,xu2025qwen2}, where we illustrate the mainstream works and the state-of-the-art MLLMs~\citep{tong2024cambrian,zhang2024videoinstructiontuningsynthetic,li2024llavaov,xu2025qwen2} in specific domains in Fig.~\ref{fig:teaser}. Impressive performance and modality breadth are contradicted in previous works, while existing open-sourced omni-modal solutions still have a large performance gap between state-of-the-art specialized LLMs, making a strong barrier between the concept of omni-modal and real-world applications. Moreover, the lack of capability in specific domains or tasks, the mass data demand, and the inadequate alignment across modalities show the suboptimal for existing omni-modal understanding models.

In this paper, we propose the \textit{Ola} model, exploring a comprehensive solution for building an omni-modal Large Language Model with comparable performance to state-of-the-art specific LLMs. We first put efforts into enhancing fundamental comprehension capabilities across multimodal domains, which we address through two key aspects: architecture design and data optimization. 1) The \textit{Ola} framework features an extendable yet concise architecture supporting omni-modal inputs. Specifically, we develop visual and audio encoders capable of processing diverse inputs, including images, videos, speech, and music. To enable effective cross-modal integration, we design a joint alignment module incorporating local-global attention pooling for visual inputs, while enabling flexible combinations of visual, audio, and text tokens. 2) We systematically curate high-quality multimodal datasets to achieve tier-1 benchmark performance. This includes designing vision-language pre-training strategies that enhance both ViT and LLMs, and we meticulously assemble supervised fine-tuning datasets for both visual and audio modalities.

Based on a strong multi-modal understanding model, we further explore the techniques for combining all the modalities towards a comprehensive omni-modal language model without performance drop. Based on our observation across modalities, we find that video can act as the core for bridging all the modalities, as visual, audio, and video subtitles show a high correlation in one sample. Therefore, we can deeply excavate the relationships between video and the corresponding audio to construct the bridge between visual and audio modality. Specifically, we collect videos from academic and open-ended web sources, design separate cleaning pipelines, and then utilize vision-language models to generate question-answering pairs based on the subtitles and video content. To operate omni-modal training, we design the progressive modality alignment strategy, making omni-modal learning easier by disassembling the complex training procedure into small steps, therefore maintaining a small size of cross-modal alignment data and making it easier to start from existing achievements in vision-language models. As shown in Fig.~\ref{fig:compare} (c), the performance of Ola largely benefits from our progressive training pipeline, leading to more balanced and competitive results on all modalities.

We evaluate \textit{Ola} under the complete omni-modal benchmarks, including image, video, and audio aspects. With only 7B parameters, \textit{Ola} achieves competitive performance across mainstream multi-modal benchmarks. On Image Benchmarks, \textit{Ola} excels at general understanding and specific-task understanding, with an overall mean accuracy of 72.6\% on the challenging OpenCompass benchmark~\citep{duan2024vlmevalkit}, 84.3\% average scores on MMBench-1.1~\citep{liu2023mmbench}, 57.0\% average scores on MMMU~\citep{yue2024mmmu}, etc. On the challenging VideoMME~\citep{fu2024videomme} multiple-choice benchmark ranging from videos within 30 seconds to 1 hour, \textit{Ola} achieves the impressive accuracy of 68.4\% with video and audio inputs. \textit{Ola} also excels at audio understanding tasks such as audio-speech recognition and chat evaluation, achieving 3.1 mean WER on LibriSpeech~\citep{panayotov2015librispeech} and 6.41 GPT-eval score on AIR-Bench~\citep{yang2024airbench}. Results on the benchmarks show a giant promotion compared with existing omni-modal LLMs and even outperforming the performance of state-of-the-art specialized LLMs.

\section{Related Works}
\label{sec:related_works}

\paragrapha{Large Vision-Language Models. }Inspired by the success of AI assistants and large language models~\citep{GPT35,GPT4V,OpenAI_GPT4_2023}, research has increasingly focused on vision-language multi-modal large language models. Significant advancements have been made in architecture design~\citep{alayrac2022flamingo,dong2024insight,liu2024llava,liu2024efficient}, training strategies~\citep{chen2024internvl,liu2024chain}, model scaling~\citep{li2024llavaov}, and data curation~\citep{laurençon2024cauldron,lu2024deepseek}. Furthermore, models are evolving beyond static images to support video~\citep{lin2023videollava,maaz2023videochatgpt,chen2024sharegpt4video}, 3D~\citep{hong20233d}, and mixed visual inputs~\citep{ranzinger2024radio,qwen2vl}. However, extending these models to effectively integrate audio modalities while maintaining balanced and robust performance remains an area that has not been fully explored.

\paragrapha{Large Audio-Text Models. }Large Language Models, mainly focused on text inputs and outputs, have a foundational link to speech, with pioneering efforts integrating speech inputs through adapter-based modifications~\citep{chen2023xllm,wu2023decoder,fathullah2024prompting}. The challenge of LLM-based speech generation has been addressed with the development of speech decoders~\citep{zhang2023speechgpt,rubenstein2023audiopalm}, marking a significant step towards omni-modal models. Beyond speech, research is expanding into audio-based LLMs that encompass the understanding of music, events, and more. Notable examples include AudioGPT~\citep{huang2024audiogpt} and SALMONN~\citep{tang2023salmonn}, which explore these audio dimensions, while models like Qwen2-Audio~\citep{chu2024qwen2audio} demonstrate enhanced understanding capabilities.

\paragrapha{Towards Large Omni-Modal Models. }Recent advancements in large language models~\citep{GPT4o,reid2024gemini} have spurred interest in developing omni-modal models that can handle multiple modalities simultaneously.  Notable examples include SpeechGPT~\citep{zhang2023speechgpt} and LLaMA-Omni~\citep{fang2024llamaomni}, which integrate audio-text understanding with speech generation. VITA~\citep{fu2024vita,fu2025vita1_5} and Qwen2.5-Omni~\citep{xu2025qwen2} extends this capability by unifying audio, image, video, and text understanding. However, existing omni-modal models often fall short in managing the full spectrum of input modalities and output formats, or they suffer from significantly poorer performance.  Ola aims to address these limitations by enhancing the capability and efficiency of omni-modal models with better architecture, training strategy, and data preparation.

\section{\emph{\color{Orange}Ola}: Omni-Modal Understanding}
\label{sec:method}

\begin{figure*}[t]
\centering
\includegraphics[width=0.95\textwidth]{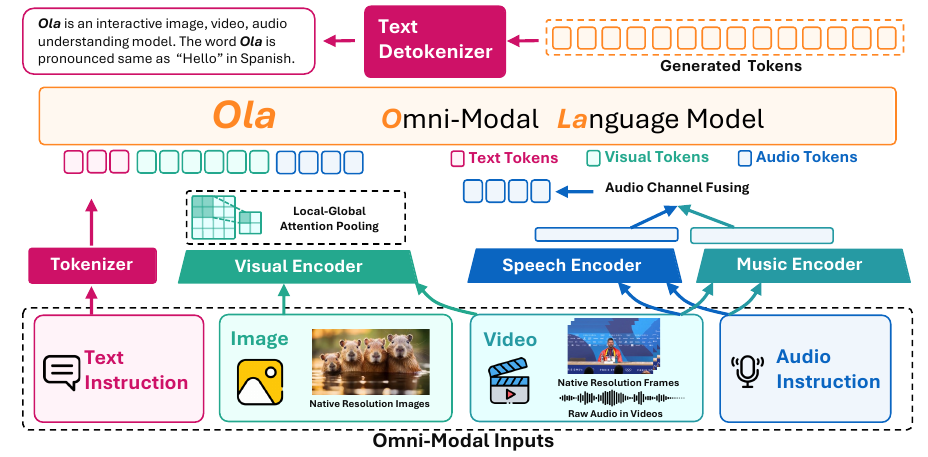} \vspace{-5pt}
\caption{\small  \textbf{\textit{Ola} Architecture. }\textit{Ola} supports omni-modal inputs including text, image, video, and audio, capable of processing the inputs simultaneously with competitive performance on understanding tasks.}
\label{fig:architecture}\vspace{-10pt}
\end{figure*}

In this section, we present our innovations in building a comprehensive multimodal framework capable of processing arbitrary visual, auditory, and textual inputs. We illustrate the architecture of \textit{Ola} in Fig.~\ref{fig:architecture}. Building upon a pre-trained Large Language Model foundation, we first elaborate our designs for visual content processing in Section~\ref{sec:method_visual} and audio content integration in Section~\ref{sec:method_audio}. We further conduct an in-depth investigation of modality alignment challenges in omni-modal learning through Section~\ref{sec:method_modality}. To enable effective cross-modal learning, we establish video as the central medium by constructing cross-modal video datasets that emphasize audio-visual correlation learning. Moreover, we design the progressive training strategy to bridge the modality gaps between language and vision from primary to periphery.

% \begin{figure*}[t]
% \centering
% \includegraphics[width=1\textwidth]{figs/fig1-new.pdf} \vspace{-15pt}
% \caption{\textbf{Illustrations of the \textit{Ola} Progressive Modality Alignment. }We visualize the relationships among modalities in the left part. Speech acts as the connection between language and audio knowledge, while video constructs the bridge with highly relevant visual and audio information. Therefore, we design the progressive alignment training strategy from primary to periphery. Furthermore, we design the cross-modality video-audio data to better capture the relationships among modalities. }
% \label{fig:training} \vspace{-15pt}
% \end{figure*}

\subsection{Advanced Visual Understanding}
\label{sec:method_visual}

As illustrated in Fig.~\ref{fig:compare} (a), starting from a fundamental vision-language model~\citep{li2024llavaov}, we present a high-quality and state-of-the-art visual understanding model through our improvements on architectures, pre-training, and fine-tuning data.

\paragrapha{Visual Encoding.} For visual inputs that include images $I$, videos $V_{1, 2, \cdots, n}$ with $n$ frames, the visual encoding part transforms the pixels into embeddings at language spaces. In the \textit{Ola} visual encoding, we preserve the original aspect ratio of each image or frame for the arbitrary resolution vision encoder OryxViT~\citep{liu2024oryx} initialized from SigLIP-400M~\citep{zhai2023siglip}, as OryxViT performs a more natural solution for visual inputs. With multi-modal visual encoder $\mathcal{E}_{I,V}(I, V_{f_1, f_2, \cdots})$ for encoding, we obtain the image features $f_{I},$ and the video features for each frame $[f_{V_{1}},f_{V_{2}}, \cdots, f_{V_{n}}]$ based on the image patches.

The alignment modules act as the converter from specific-modal spaces to the text embedding space. \textit{Ola} treats images and video frames equally to hold the unification of visual contents. To reduce the token length of visual features for higher efficiency, we take one step forward based on the motivation of structural downsampling in previous works~\citep{li2024minigemini}, and propose the \textit{Local-Global Attention Pooling} layer for better downsampled features with less information loss. Specifically, for image or video frame feature in spatial shape $H\times W$ and channel $C$, we adopt the bilinear interpolation for 2x downsampling to obtain $f^{\text{global}}$, which contains the global information of the downsampled region. We combine the original and global features for local-global embeddings and use Softmax to predict the importance $\pi$ of each downsampled spatial region:
\begin{equation}
    f=\text{Concat}[f,f^{\text{global}}], \ \ \pi=\text{Softmax}(\text{MLP}(f))
\end{equation}
The downsampled feature $f^{\text{global}}$ determines the weight of each previous region based on the score $\pi$ with the Hadamard product, therefore assigning more weights to the more informative part during the downsampling process.

\paragrapha{Visual Pre-Training at Scale. }We approach scalable visual pre-training through two key components: ViT pre-training and instruction pre-training. The ViT pre-training is independently conducted on vision transformers, integrated with a small LLM as a language interface. The primary objective of this phase is to enhance the visual-language capabilities of the visual encoder. To achieve this, we utilize large-scale data pairs derived from OCR, grounding, and captioning datasets. The subsequent instruction pre-training stage aims to equip the \textit{Ola} model with comprehensive knowledge. We collect approximately 20M text-image pairs from both open-source and in-house sources to establish foundational capabilities. To ensure data quality, state-of-the-art vision-language models are employed for re-captioning and re-questioning of the data pairs, ultimately generating a substantial volume of high-quality, instruction-level pre-training data. The detailed data composition and modification pipeline are stated in the Appendix.

\paragrapha{Supervised Fine-Tuning. }The supervised fine-tuning visual data is collected from academic datasets in image and video categories. For the image data, we follow the simple setting in~\citep{liu2024llava15} for image MLP alignment. The MLP alignment data includes 800k image captioning pairs from the LAION dataset~\citep{schuhmann2021laion}. For text-image supervised fine-tuning data, we collect abundant data from various tasks, including captions, conversations, OCR, etc. The source of the training data involves the mixture of LLaVA-OneVision~\citep{li2024llavaov}, Cauldron~\citep{laurençon2024cauldron}, Cambrian-1~\citep{tong2024cambrian}, MAmmoTH-VL~\citep{guo2024mammoth}, PixMo~\citep{deitke2024molmo}, etc., resulting in 7.3M image training data in total. 

For text-video training data, we collect useful video datasets from LLaVA-Video-178K~\citep{zhang2024videoinstructiontuningsynthetic}, VideoChatGPT-Plus~\citep{maaz2023videochatgpt}, LLaVA-Hound~\citep{zhang2024hound}, and Cinepile~\citep{rawal2024cinepile}, with 1.9M video conversation pieces in total. We randomly sample 2/3 video-language data pairs from LLaVA-Video-178K, resulting in 1.2M high-quality training data, and we use the full set of other data sources.

\begin{figure*}[t]
\centering
\includegraphics[width=1\textwidth]{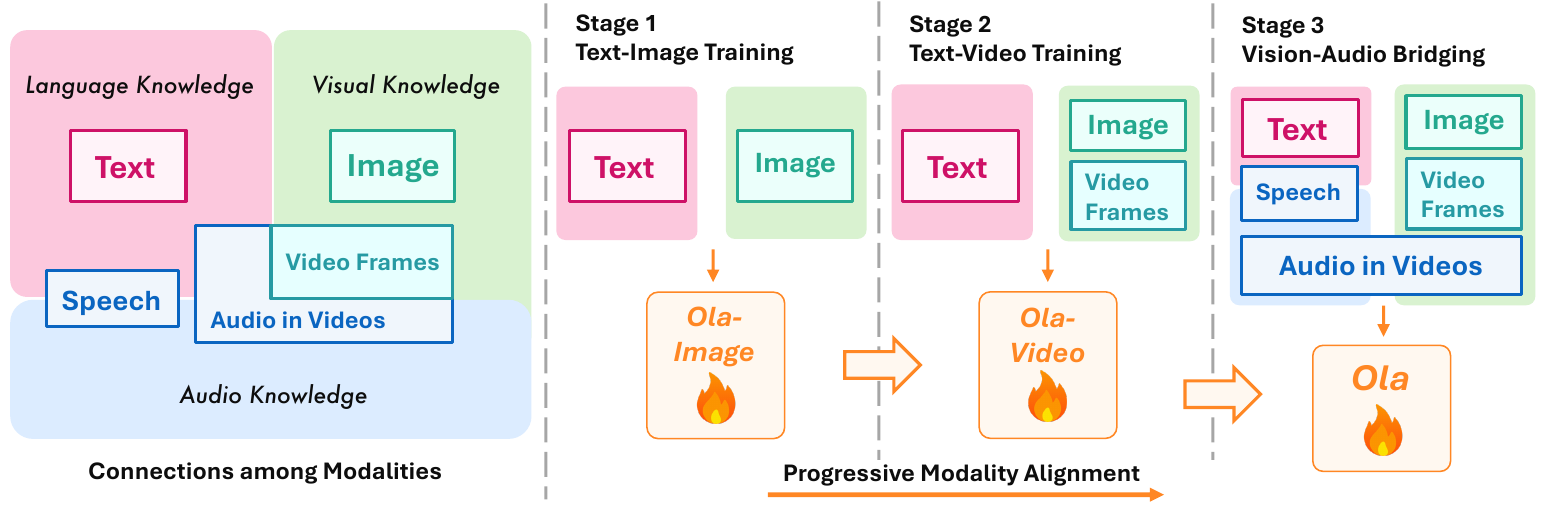}
% \fbox{\rule{0pt}{2.5in} \rule{0.9\linewidth}{0pt}}
\caption{\small \textbf{Illustrations of the Relations between Modalities.} We visualize the relationships among modalities in the left part. Speech acts as the connection between language and audio knowledge, while video constructs the bridge with highly relevant visual and audio information. Therefore, we design the cross-modality video-audio data to better capture the relationships among modalities. Furthermore, we design the progressive alignment training strategy from primary to periphery.}
\label{fig:training} 
\end{figure*}

\subsection{Robust Audio Integration}
\label{sec:method_audio}

The audio encoding component serves as a critical element in omni-modal models, as the system fundamentally relies on comprehending complex auditory inputs encompassing user speech signals and multimodal audio-visual comprehension data. In the \textit{Ola} model, we implement a robust audio recognition module capable of processing speech, music, and video content, which seamlessly integrates with visual processing components.

\paragrapha{Dual Audio Encoding. }For audio encoding, we propose the dual encoder approach for the audio input. Specifically, we use Whisper-v3~\citep{radford2022whisper} as the speech encoder and BEATs~\citep{Chen2022beats} as the music encoder for better alignment between audios and texts, providing richer audio information. The music encoder takes the original wav audio $A$ as inputs, while the speech encoder takes the wav transformed into Mel spectrogram representation $A_{(mel)}$ as inputs. Note that the Whisper encoder only supports a certain length of audio inputs, therefore we fix the sample rate as 16000Hz and cut the overlong audio into segments of 30 seconds $A_1, A_2, \cdots, A_n$ and conduct the encoder operation in batches $[f_{A_1},f_{A_2},\cdots,f_{A_n}]=\mathcal{E}_A([A_1, A_2, \cdots, A_n])$. The embedding features of the speech and music encoders are concatenated across channel dimensions for the comprehensive audio features $f_A$. 

We apply the simple yet effective two-layer non-linear MLP connectors $\text{MLP}_{A},\text{MLP}_{V}$ following the previous works~\citep{liu2024llava,liu2024llava15} to project the specific modal features $[f_I,f_V,f_A]$ into unified tokens $[t_I,t_V,t_A]$. We define visual and audio start, separate, newline, and end tokens to indicate special positions for inputs. The omni-modal tokens $[t_I,t_V,t_A]$ are concatenated with text tokens $t_T$ in free combination for LLM decoding. 

\paragrapha{Mixing Audio Data. }The audio training data are collected from various audio-relevant environments, including comprehensive speech and music understanding. For text-speech understanding, we design multiple tasks including ASR from LibriSpeech~\citep{panayotov2015librispeech} and GigaSpeech~\citep{chen2021gigaspeech} datasets, audio captioning from AudioCaps~\citep{kim2019audiocaps} and Clotho~\citep{drossos2020clotho} datasets, speech question answering from LibriSpeech~\citep{panayotov2015librispeech} datasets, audio question answering from WavCaps~\citep{mei2024wavcaps} and AudioCaps~\citep{kim2019audiocaps} datasets. For text-music understanding, we collect tasks including music captioning from MusicCaps~\citep{agostinelli2023musiclm}, music question answering from MillionSong~\citep{mcfee2012millionsong} and MusicNet~\citep{thickstun2016musicnet}. The overall audio training data involves 1.1M samples. The relevant text question-answering representations are collected from SALMONN~\citep{tang2023salmonn}.

\subsection{Bridging Modality Gaps for Omni-Modal Understanding}
\label{sec:method_modality}

\paragrapha{Rethinking Modality Gaps among Language, Vision and Audio. }From our exploration, we recognize two critical issues in omni-modal training. 1) \textit{Connections between modalities.} In conventional learning strategies for omni-modal models, the training samples emphasizing the connections between modalities are always lacking, especially in two major modalities: vision and audio. From our efforts, we observe that jointly learning audio and vision data can yield surprising results in omni-modal learning by providing a more comprehensive view across different modalities. For the \textit{Ola} model, we consider video as the bridge between audio and vision, as videos contain natural, abundant, and highly relevant information between frames and the accompanying audio. We test our hypothesis by optimizing the training pipeline and preparing targeted training data, as introduced below. 2) \textit{Modal balancing.} As illustrated in Fig.~\ref{fig:compare} (c), directly combining data from all modalities negatively affects benchmark performance. Therefore, we propose a rational training procedure that progressively equips the sense organ to the \textit{Ola} model. We assert that texts and images are the core modalities in omni-modal learning, while speeches and videos are variants of texts and images, respectively. Learning to recognize texts and images ensures the model's basic cross-modal ability, so we prioritize these harder cases. Subsequently, we gradually incorporate video, audio, and speech into the training for the omni-modal LLM. 

\subsubsection{Deriving Omni-Modal Model from Videos}

Most existing video training data are annotated or synthesized solely from frame inputs, often overlooking the valuable information in accompanying audio. To address this, we designed a pipeline to generate cross-modal video data, aiming to uncover the intrinsic relationships between video and audio. This guides an omni-modal large language model in learning cross-modality information. Specifically, we developed two tasks for cross-modal learning: video-audio question answering and video speech recognition. We collected videos from the academic dataset LLaVA-Video-178k~\citep{zhang2024videoinstructiontuningsynthetic} and the open-ended video datasets from FineVideo~\citep{Farré2024FineVideo}. Due to the lack of subtitles in the academic datasets, we used Whisper-v3~\citep{radford2022whisper} to generate subtitles from the video audio and conducted a language-based cleaning procedure. We then employed a large language model to assess whether the subtitles were complete and informative. We gathered 41k pure videos from LLaVA-Video-178k, along with the original 42k videos from FineVideo. Subsequently, we used Qwen2-VL-72B~\citep{qwen2vl} to generate questions and answers based on the video and corresponding subtitles. The model was instructed to focus on the subtitle inputs while using the videos as supplementary information. We created three question-answer pairs for each video, resulting in 243k cross-modal video-audio data. Additionally, we included the original video subtitling tasks with 83k training data to help the model maintain its ASR ability in noisy environments. During training, the models processed multiple frames, audio, and text inputs simultaneously, significantly enhancing their cross-modal capabilities.

\subsubsection{Omni-Modal Training with Progressive Modality Alignment}

We start from two fundamental and separated modalities, image and text, to build the basic knowledge for omni-modal models. Subsequently, we gradually expand the training sets to equip the model with an extended ability, including video frames that strengthen the visual understanding capability, speech data that connects the language and audio knowledge, and the video with audio that mixes up the information from language, video, and audio comprehensively. 

\paragrapha{Stage 1: Text-Image Training. }The \textit{Ola} training starts from a pre-trained Large Language Model, where we use Qwen2.5-7B~\citep{qwen2.5} in our implementation for better trade-offs for model sizes and performance. The \textit{Ola} text-image training includes MLP alignment, large-scale pre-training, and supervised fine-tuning following common practice in large-scale multi-modal learning~\citep{tong2024cambrian,li2024llavaov}. We initialize the visual MLP adapter and freeze other parameters in MLP alignment with the image captioning task. Subsequently, we unfreeze all the parameters including the vision encoder in the pre-training and supervised fine-tuning phase. The downsampling module is well-trained in the text-image training stage to hold the 2x compression for visual data including images and videos.

\paragrapha{Stage 2: Continuous Training for Images and Videos. }Based on a strong text-image multi-modal LLM, we continuously extend the capability for \textit{Ola} with video data. We keep most of the experimental setting for supervised fine-tuning while freezing the vision encoder in this stage as the encoder is already fully trained beforehand. We mix the previous image data and the video data to preserve the original text-image performance. We randomly sample 0.8M image data from stage 1 and mix it with the video datasets for continuous training to maintain the basic performance. 

\paragrapha{Stage 3: Bridging Vision and Audio with Videos. }The audio-relevant training is included in stage 3. We follow the training strategy for the visual MLP adapter while initializing the audio MLP adapter with a basic audio-speech recognition (ASR) task. Then we mix up the text \& speech understanding, text \& music understanding, audio \& video joint comprehension, and the foremost text-image multi-modal tasks together for the formal training. \textit{Ola} concentrates on learning audio recognition and identifying the relationships between vision and audio in this stage, resulting in a comprehensive image, video, and audio understanding model. We mix up all the 324k cross-modal data, 1.1M pure text-audio data for the comprehensive training stage. Additionally, we sample 400k image data from stage 1 to maintain the basic ability and create 200k image data with voice instructions to equip the model with interaction capability.

\section{Experiments}
\label{sec:exp}

\begin{table*}[t]
  \centering
  \caption{\small \textbf{Main Results across Image, Video, and Audio Understanding Benchmarks. }We select representative benchmarks among image, video, and audio benchmarks, and select mainstream state-of-the-art open-source large language models in each modality. We also include open-source omni-modal LLMs for comparison. In the table, "$-$" indicates the model is capable of solving the tasks theoretically, while the result is lacking. "\xmark" indicates that the model is not capable of the task. $\downarrow$ indicates that lower score is better. * LLaMA-Omni is not optimized for ASR and thus cannot produce reasonable results on this task.} \vspace{-5pt}
  \adjustbox{width=\linewidth}{
    \begin{tabular}{llcccccccccccc}
    \toprule
    \multirow{2}[2]{*}{Model} & \multirow{2}[2]{*}{Size} & \multicolumn{7}{c}{Image Benchmarks} & \multicolumn{3}{c}{Video Benchmarks} & \multicolumn{2}{c}{Audio Benchmarks} \\
    \cmidrule(lr){3-9}\cmidrule(lr){10-12}\cmidrule(lr){13-14}
          &       & \rotatebox{45}{MMBench-1.1} & \rotatebox{45}{MMStar} & \rotatebox{45}{MMMU} & \rotatebox{45}{MathVista} & \rotatebox{45}{HalluBench} & \rotatebox{45}{AI2D} & \rotatebox{45}{OCRBench} & \rotatebox{45}{VideoMME} & \rotatebox{45}{LongVideoBench} & \rotatebox{45}{MVBench} & \rotatebox{45}{LibriSpeech$\downarrow$} & \rotatebox{45}{AIR-Bench} \\
    \midrule
    \multicolumn{14}{l}{\textit{Image LLMs}} \\
    \midrule
    Cambrian-1~\citep{tong2024cambrian} & 8B & 68.2 & 50.7 & 41.8 & 48.1 & 30.6 & 74.6 & 614 & \xmark & \xmark & \xmark & \xmark & \xmark \\
    Pixtral~\citep{agrawal2024pixtral} & 12B & 72.7 & 54.5 & 51.1 & 56.3& 47.0 & 79.0 & 685 & \xmark & \xmark & \xmark & \xmark & \xmark \\
    \midrule
    \multicolumn{14}{l}{\textit{Video LLMs}} \\
    \midrule
    VideoCCAM~\citep{fei2024videoccam} & 9B   &$-$  &$-$   &$-$  &$-$   &$-$    &$-$    &$-$    & 53.9  &$-$    & 64.6  & \xmark & \xmark \\
    LLaVA-Video~\citep{zhang2024videoinstructiontuningsynthetic} & 7B  &$-$  &$-$  &$-$    &$-$  &$-$   &$-$    &$-$ & 63.3 & 58.2 & 58.6  & \xmark & \xmark \\
    \midrule
    \multicolumn{14}{l}{\textit{Vision Comprehensive LLMs}} \\
    \midrule
    LLaVA-OneVision~\citep{li2024llavaov} & 7B & 80.9 & 61.9 & 47.9 & 62.6 & 31.6 & 82.4 & 622 & 58.2  & 61.3  & 59.4  & \xmark & \xmark \\
    MiniCPM-V 2.6~\citep{yao2024minicpm} & 8B & 78.0 & 57.5 & 49.8 & 60.8 & 48.1 & 82.1 & 852 & 60.9  &$-$    &$-$    & \xmark & \xmark \\
    %Qwen2-VL~\citep{qwen2vl} & 7B & 81.0 & 60.7 & 53.7 & 61.6 & 50.4 & 83.0 & 843 & 63.3 & $-$ &$-$    & \xmark & \xmark  \\
    InternVL2.5~\citep{chen2024intern25} & 8B  & 82.5 & 63.2 & 56.2 & 64.5 & 49.0 & 84.6 & 821 & 64.2 & 60.0 & \textbf{72.0}   & \xmark & \xmark   \\
    Qwen2.5-VL~\citep{Qwen2.5-VL} & 7B & 82.6 & 64.1 & 56.2 & 65.8 & \textbf{56.3} & 84.1 & \textbf{877} & 65.1 & 54.7 & 69.6 &  \xmark & \xmark \\
    \midrule
    \multicolumn{14}{l}{\textit{Audio LLMs}} \\
    \midrule
    SALMONN~\citep{tang2023salmonn} & 13B  & \xmark & \xmark  & \xmark & \xmark & \xmark & \xmark & \xmark & \xmark & \xmark & \xmark & 3.5   & 6.12  \\
    Qwen2-Audio~\citep{chu2024qwen2audio} & 7B   & \xmark & \xmark & \xmark  & \xmark & \xmark & \xmark & \xmark & \xmark & \xmark & \xmark & \textbf{2.5}   & \textbf{6.93}  \\
    \midrule
    \multicolumn{14}{l}{\textit{Omni-Modal LLMs}} \\
    \midrule
    LLaMA-Omni~\citep{fang2024llamaomni} & 8B & \xmark & \xmark & \xmark & \xmark & \xmark & \xmark & \xmark & \xmark & \xmark & \xmark & 120.4$^*$ & 4.70  \\
    Mini-Omni2~\citep{xie2024miniomni2} & 0.5B  & 32.1 &$-$ & 24.9 &$-$ &$-$ &$-$ & 6 &$-$    &$-$    &$-$    & 7.2   & 3.20  \\
    VITA-1.5~\citep{fu2025vita1_5}  & 7B & 76.8 & 60.2 & 52.6 & 66.2 & 44.6 & 79.2 & 741 & 56.1 & $-$ & 55.4 & 5.4 & 4.54 \\
    IXC2.5-OmniLive~\citep{internlmxcomposer2_5_OL} & 8B & 79.4 & 59.9 & 42.9 & 64.0 & 43.1 & 81.6 & 686 & 60.6 & $-$ & 68.7 & 4.4 & 1.67\\
    Qwen2.5-Omni~\citep{xu2025qwen2} & 7B & 81.8 & 64.0 & \textbf{59.2} & 67.9 & $-$ & 83.2 & $-$ & 64.3 & $-$ & 70.3 & 2.6 & 5.72 \\
    \midrule
    \rowcolor{Lightorange}\textit{\textbf{Ola}} & 7B & \textbf{84.3} & \textbf{70.8} & 57.0 & \textbf{68.4} & 53.5 & \textbf{86.1} & 827 & \textbf{68.4} & \textbf{61.4} & 66.3 & 3.1 & 6.41 \\
    \bottomrule
    \end{tabular}%
    }
  \label{tab:main} \vspace{-15pt}
\end{table*}%
We conduct all-sided benchmarking in Sec.~\ref{sec:exp_main}  to evaluate the all-powerful \textit{Ola} model, including the representative benchmarks in image, video, and audio understanding. Subsequently, we conduct detailed results on critical benchmarks in Sec.~\ref{sec:exp_analysis} to demonstrate the effectiveness of our design on motivation, training, and data preparations. 

\subsection{Implementation Details}

The \textit{Ola} model builds upon the Qwen-2.5-7B~\citep{qwen2.5} framework, incorporating OryxViT~\citep{liu2024oryx} as the vision encoder initialized from SigLIP-400M~\citep{zhai2023siglip}, Whisper-V3-Large~\citep{radford2022whisper} as the speech encoder, and BEATs-AS2M(cpt2)~\citep{Chen2022beats} as the music encoder. Initially, we employ a relatively high learning rate of 1e-3 for MLP adapter pre-training. During supervised fine-tuning, the learning rate is gradually reduced from 2e-5 for text-image and multi-image training to 1e-5 for video-audio training. We utilize a batch size of 256 for fine-tuning, leveraging 64 NVIDIA A800 GPUs to conduct our training. We adopt 10$\times$ downsampled rate for audio features to reduce the token length, resulting in 300 tokens per minute. During training and inference, the maximum token length is set to 16384 and the maximum number of audio trunks is set to 25.

\subsection{Results on Omni Understanding} \label{sec:exp_main}

\paragrapha{Benchmarks. }We conduct extensive comparisons across image, video, and audio understanding benchmarks to demonstrate the omni-modal capabilities of the \textit{Ola} model. For image benchmarks, we utilize comprehensive understanding datasets including MMBench-1.1~\citep{liu2023mmbench}, MMMU~\citep{yue2024mmmu}, MMStar~\citep{chen2024mmstar}, MathVista~\citep{lu2023mathvista}, Hallusion Bench~\citep{guan2024hallusionbench}, AI2D~\citep{kembhavi2016ai2d}, and OCRBench~\citep{liu2023ocrbench}. In the video domain, we evaluate using VideoMME~\citep{fu2024videomme}, which involves multiple-choice questions on videos of varying lengths, and LongVideoBench~\citep{wu2024longvideobench} for assessing performance on extremely long video content, MVBench~\citep{li2024mvbench} for the general recognition ability. For audio benchmarks, we focus on two primary tasks relevant to audio LLMs. Librispeech~\citep{panayotov2015librispeech} serves as a traditional audio-speech recognition (ASR) dataset, testing the model's ability to accurately transcribe spoken language. AIR-Bench~\citep{yang2024airbench} provides a comprehensive evaluation of audio question-answering capabilities, incorporating speech, sound, and music inputs. The responses are evaluated with a GPT-based~\citep{OpenAI_GPT4_2023} scorer against ground truth answers. We report the mainstream evaluation metric on image and video benchmarks and report the mean metric for audio understanding tasks for simplicity. 

\paragrapha{Baselines. }We selecte a range of state-of-the-art MLLMs across different modalities for comparison and reference. We categorize vision-language models into three groups: image-centric LLMs, video-centric LLMs, and comprehensive LLMs capable of handling both images and videos.  For image understanding, we utilized Cambrian-1~\citep{tong2024cambrian} and Pixtral-12B~\citep{agrawal2024pixtral}.  For video understanding, VideoCCAM~\citep{fei2024videoccam} and LLaVA-Video~\citep{zhang2024videoinstructiontuningsynthetic} are employed. Comprehensive models included LLaVA-OneVision~\citep{li2024llavaov}, MiniCPM-V 2.6~\citep{yao2024minicpm}, InternVL2.5~\citep{chen2024intern25}, and Qwen2.5-VL~\citep{Qwen2.5-VL} which excel across various visual benchmarks. In the audio domain, we compared our work with SOTA models such as SALMONN~\citep{tang2023salmonn} and Qwen-2 Audio~\citep{chu2024qwen2audio}. As an Omni-modal LLM, our model, \textit{Ola}, was compared with SOTA open-source omni-modal LLMs like Mini-Omni2~\citep{xie2024miniomni2}, VITA-1.5~\citep{fu2024vita}, InternLM-XComposer2.5-OmniLive~\citep{internlmxcomposer2_5_OL}, Qwen2.5-Omni~\citep{xu2025qwen2}.  Additionally, LLaMA-Omni~\citep{xie2024miniomni2}, an audio-text omni model, was noted for its strong speech generation capabilities.

\begin{table*}[t]
  \centering
  \caption{\small \textbf{Analysis Results on Audio Benchmarks. }We report the WER rate on test-clean, test-other, dev-clean, dev-other subsets of LibriSpeech, the scores on AIR-Bench, MMAU, and AIR-Foundation.. In the table, "$-$" indicates the model is capable of solving the tasks, while the result is lacking. "\xmark" indicates that the model is not capable of the task.}
  \adjustbox{width=\linewidth}{
    \begin{tabular}{lcccccccccccccc}
    \toprule
    \multirow{2}[2]{*}{Model} & \multicolumn{4}{c}{ASR on LibriSpeech$\downarrow$} & \multicolumn{5}{c}{Chat on AIR-Bench} & \multicolumn{1}{c}{MMAU} & \multicolumn{4}{c}{AIR-Foundation} \\
    \cmidrule(lr){2-5}\cmidrule(lr){6-10}
    \cmidrule(lr){11-11}\cmidrule(lr){12-15}
          & ~test-c~ & ~test-o~ & ~dev-c~ & ~dev-o~ & ~speech~ & ~sound~ & ~music~ & ~mix~ & ~avg~ & ~~testmini~~ & ~speech~ & ~sound~ & ~music~ & ~avg~ \\
    \midrule
    \textit{Audio Models} \\
    \midrule
    SpeechGPT~\citep{zhang2023speechgpt} & \xmark & \xmark & \xmark & \xmark & 1.6  & 1.0  & 1.0  & 4.1 & 1.9 &  $-$  & 34.3 & 27.5 & 28.1 & 30.0\\
    Whisper-small~\citep{radford2022whisper} & 4.4   & 10.1  & 4.6   & 10.3  & \xmark & \xmark & \xmark & \xmark & \xmark & \xmark & \xmark & \xmark & \xmark & \xmark \\
    SALMONN~\citep{tang2023salmonn} & 2.1   & 4.9   &$-$    &$-$    & 6.2  & 6.3  & 6.0  & 6.1 & 6.1 & 33.7  & 37.8 & 33.0 & 37.1 & 36.0\\
    Qwen2-Audio~\citep{chu2024qwen2audio} & 1.6   & 3.6   & 1.3   & 3.4   & 7.2  & 7.0  & 6.8  & 6.8 & 6.9 & 49.2 & 62.9 & 55.4 & 56.8 & 58.4\\
    \midrule
    \textit{Omni-Modal LLMs} \\
    \midrule
    LLaMA-Omni~\citep{fang2024llamaomni} & \xmark & \xmark & \xmark & \xmark & 5.2  & 5.3  & 4.3  & 4.0 & 4.7 & $-$ & $-$ & $-$ & $-$ & $-$ \\
    Mini-Omni2~\citep{xie2024miniomni2} & 4.8   & 9.8   & 4.7   & 9.4   & 3.6  & 3.5  & 2.6  & 3.1 & 3.2 &$-$  &$-$ &$-$ &$-$ &$-$\\
    VITA-1.5~\citep{fu2025vita1_5} & 3.3 & 7.2 & 3.4 & 7.5 & 4.8 & 5.5 & 4.9 & 2.9 & 4.5 & 35.5  & 31.5 & 24.1 & 25.5 & 27.0\\
    IXC2.5-OmniLive~\citep{internlmxcomposer2_5_OL} & 2.5 & 5.7 & 2.6 & 5.8 & 1.6 & 1.8 & 1.7 & 1.6 & 1.7 &$-$ &$-$ &$-$ &$-$ &$-$  \\
    Qwen2.5-Omni~\citep{xu2025qwen2} & \textbf{1.6} & \textbf{3.5} & \textbf{1.8} & \textbf{3.4} & 6.8 & 5.7 & 4.8 & 5.4 & 5.7 & 65.6 & \textbf{67.2} & \textbf{76.3} & \textbf{63.0} & \textbf{68.8}\\
    \midrule
    \textbf{\textit{Ola}} (Pure audio) & 2.1 & 4.7 & 2.1 & 4.6 & 6.3 & 5.4 & 5.8 & 5.8  & 5.8 & 66.2  & 55.7 & 67.6 & 50.4 & 57.9\\
    \rowcolor{Lightorange}\textbf{\textit{Ola}} & 1.9 & 4.4 & 1.9 & 4.2 & \textbf{7.3} & \textbf{6.4} & \textbf{5.9} & \textbf{6.0} & \textbf{6.4} & \textbf{70.3} & 58.8 & 70.4 & 53.1 & 60.8\\
    \bottomrule
    \end{tabular}%
    }
  \label{tab:audio}
\end{table*}%

\paragrapha{Results. }We present the results in Table~\ref{tab:main}, highlighting \textit{Ola}'s competitive performance across major multi-modal benchmarks when compared to state-of-the-art specialist-modal LLMs. Specifically, in image benchmarks, \emph{Ola} achieves 84.3\% on MMBench-1.1~\citep{liu2023mmbench}, 70.8\% on MMStar~\citep{chen2024mmstar}, 57.0\% on MMMU~\citep{yue2024mmmu}, 68.4\% on MathVista~\citep{lu2023mathvista}, 86.1\% on AI2D~\citep{kembhavi2016ai2d} and 827 on OCRBench~\citep{liu2023ocrbench}, surpassing all the relative multi-modal LLMs in similar number of parameters. In video benchmarks, \textit{Ola} attains an impressive 68.4\% on VideoMME~\citep{fu2024videomme}, showcasing its robust capability to handle both video and audio inputs simultaneously, and setting a new state-of-the-art performance among 7B models on the VideoMME benchmark. \emph{Ola} also maintains a leading position compared to mainstream video LLMs including LLaVA-Video~\citep{zhang2024videoinstructiontuningsynthetic} and VideoCCAM~\citep{fei2024videoccam} on LongVideoBench~\citep{wu2024longvideobench} and MVBench~\citep{li2024mvbench}. In audio benchmarks, \textit{Ola} demonstrates strong audio-speech recognition and conversational abilities, with 3.1\% mean WER rate on LibriSpeech~\citep{panayotov2015librispeech} and 6.41 mean score on AIR-Bench~\citep{yang2024airbench}, outperforming existing omni-modal LLMs, including LLaMA-Omni~\citep{fang2024llamaomni}, which focuses on audio understanding.

\subsection{Analysis} \label{sec:exp_analysis}

For the analysis part, we report the detailed results on audio benchmarks to illustrate the fine-grained performance. We also demonstrate our designs on training and cross-modal training data with ablations on critical benchmarks. At last, we perform qualitative showcases of \textit{Ola}.

% Table generated by Excel2LaTeX from sheet 'Sheet4'
\begin{table}[t]
  \centering
  \caption{\textbf{Analysis on Omni-Modal Training. }We conduct analysis for the performance before/after omni-modal learning, and show the performance gain with audio inputs on mainstream video benchmarks. The highlighted row indicates the final accepted strategy. }\vspace{-5pt}
  \adjustbox{width=\linewidth}{
    \begin{tabular}{ccccccc}
    \toprule
    Omni-Stage & \multirow{2}[4]{*}{~Audio~} & \multirow{2}[4]{*}{Subtitle} & \multicolumn{4}{c}{VideoMME} \\
\cmidrule{4-7}  Training  &       &       & ~~Short~~ & ~~Medium~~ & ~~Long~~  & ~~Overall~~  \\
    \midrule
    \xmark & \xmark & \xmark &  75.9 & 61.2 & 54.3 & 63.8 \\
    \cmark & \xmark & \xmark & 76.4  & 61.9  & 54.8  & 64.4   \\
    \cmark & \xmark & \cmark & 78.4  & 66.6  & 56.4  & 67.1   \\
    \rowcolor{Lightorange} \cmark & \cmark & \xmark & 78.7  & 68.3  & 58.3  & 68.4 \\
    \cmark & \cmark & \cmark & 78.8  & 68.8  & 60.3  & 69.3 \\
    \bottomrule
    \end{tabular}%
    }
  \label{tab:video} \vspace{-5pt}
\end{table}%

% Table generated by Excel2LaTeX from sheet 'Sheet4'
\begin{table}[t]
  \centering
  \caption{\textbf{Analysis on Progressive Modality Learning.} We evaluate the basic performance on image and video understanding for the intermediate models during the training stage. The highlighted row indicates the final accepted strategy.} \vspace{-5pt}
  \adjustbox{width=\linewidth}{
    \begin{tabular}{ccccccc}
    \toprule
    Stage1 & Stage2 & Stage3 & MMBench-1.1 & MMMU & OCRBench & VideoMME \\
    \midrule
    \cmark & \xmark & \xmark & 83.5  & \textbf{57.5} & \textbf{832}  & \xmark \\
    \cmark & \cmark & \xmark & 83.8 & 57.2 & 820  & 63.8 \\
    \rowcolor{Lightorange} \cmark & \cmark & \cmark & \textbf{84.3} & 57.0 & 827 & \textbf{68.4} \\
    \bottomrule
    \end{tabular}%
    }
  \label{tab:training} \vspace{-10pt}
\end{table}%

\paragrapha{Analysis on Audio Benchmarks. }To demonstrate the effectiveness of our approach on audio and speech tasks, we conducted comprehensive experiments using the LibriSpeech~\citep{panayotov2015librispeech}, AIR-Bench~\citep{yang2024airbench} and MMAU~\citep{sakshi2024mmau} datasets, and we illustrate our results on Tab.~\ref{tab:audio}. Specifically, we report the Word Error Rate (WER) on the test-clean, test-other, dev-clean, and dev-other subsets of LibriSpeech, the GPT-4 eval scores on speech, sound, music, and mix sub-metrics in AIR-Bench, testmini results for MMAU, and GPT-4 eval scores on speech, sound and music for AIR-Foundation.  Our model, \textit{Ola}, is compared against state-of-the-art audio models and omni-modal LLMs.

Notably, \textit{Ola} demonstrates a significant advantage over existing omni-modal models, achieving a 6.4 average score on AIR-Bench, a 70.3 score on MMAU testmini dataset, and a 60.8 average score on AIR-Foundation. This is in contrast to the previous state-of-the-art omni-modal models, which achieved a 6.4 average score, a 70.3 score, and a 68.8 average score, respectively. \textit{Ola}'s performance is even approaching that of audio-specific models, highlighting its strong universality. 

Furthermore, we evaluated \textit{Ola} under two situations. \textit{Ola} (Pure audio) indicates that we omit video-audio data in stage 3 and replace the same amount of data with pure audio inputs, where we can observe a consistent performance gain with cross-modal joint learning. Despite the significant distribution gap between video audio and speech-relevant datasets, this improvement indicates the robust connections between video and speech modalities.

\paragrapha{Effectiveness of Omni-Modal Training. }In exploring the relationships between video and audio, we examined the effectiveness of omni-modal training and its impact on audio within videos, both during training and in benchmark outcomes. The analysis results are shown in Tab.~\ref{tab:video}.  By comparing results before and after omni-modal training (i.e., stage 3 of the progressive training strategy), we observed performance improvements from 63.8\% to 64.4\% on VideoMME~\citep{fu2024videomme}.  Additionally, incorporating audio modalities alongside raw video resulted in significant performance gains, increasing scores from 64.4\% to 68.4\% on VideoMME~\citep{fu2024videomme}. These findings suggest that audio contains valuable information that enhances overall recognition performance. Notably, the multiple-choice accuracy for \textit{Ola} with omni-modal training and audio inputs even surpasses the results with original text subtitles, with 68.4\% overall performance compared with 67.1\% overall performance. The results indicate that audio data may include more information beyond the original text information in certain benchmarks.

\begin{figure}[t]
\centering
\includegraphics[width=0.5\textwidth]{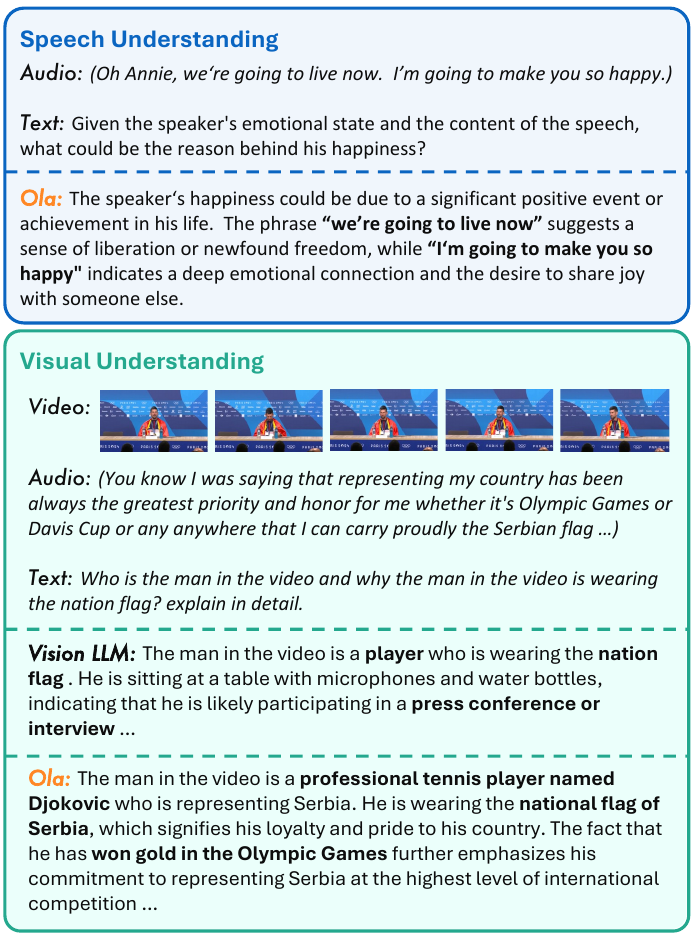} \vspace{-10pt}
\caption{\textbf{Generative results on speech and visual understanding tasks. }We show the strong ability of omni-modal \textit{Ola} compared with conventional vision-language models.}
\label{fig:showcase} \vspace{-10pt}
\end{figure}

\paragrapha{Effectiveness of Progressive Modality Learning. }To evaluate the effectiveness of the proposed training strategy, we evaluate the basic performance of the intermediate model in each stage (Stage-1 for Ola-Image, Stage-2 for Ola-Video and Stage-3 for the final \textit{Ola} model). Specifically, we adopt the representative MMBench-1.1~\citep{liu2023mmbench}, MMMU~\citep{yue2024mmmu}, and OCRBench~\citep{liu2023ocrbench} for image performance and VideoMME~\citep{fu2024videomme} for video performance. Results are shown in Tab.~\ref{tab:training}. We can observe that the progressive modality training from image, video to audio can maximally preserve the previously learned capability. Additionally, we can also observe a performance gain on MMBench-1.1 and VideoMME, revealing the superiority of joint learning. 

\paragrapha{Showcases.} We present the qualitative generation results of the \textit{Ola} model on both speech and visual understanding tasks in Fig.~\ref{fig:showcase}. For speech understanding, we utilize sources from the AIR-Bench~\citep{yang2024airbench} benchmark, where \textit{Ola} demonstrates precise speech recognition and effective emotional analysis, as well as reasoning about the questions posed. In visual understanding tasks, we analyze an interview with a famous tennis player after the Olympic Games. We compare our approach with the state-of-the-art vision-language model~\citep{li2024llavaov}. The large vision-language models without audio inputs exhibit significant information loss due to the absence of audio input and recognition capabilities for audio. In contrast, \textit{Ola}, with its omni-modal inputs, provides more accurate responses regarding nationality, context, and background from the speaker's dialogue.

\section{Conclusion} \label{sec:conclusion}

In this paper, we propose \textit{Ola}, a comprehensive and powerful omni-modal language model that achieves competitive performance in image, video, and audio understanding tasks. Our solution on architectural design, data curation, and training strategy offers a natural, efficient, and competitive pipeline for building an omni-modal model. Enhancements in architectural design with omni-modal inputs, along with high-quality pre-training and fine-tuning data preparation, extend \textit{Ola}'s capabilities. The training strategy based on joint modality learning and progressive modality alignment connects all the modalities uniformly. We hope our work inspires future research on more general AI models.

\appendix
\section*{Appendix}

We provide supplementary documents to support our research. In the appendix, we provide more details, including model details, training details, and data details in Sec.~\ref{appendix:A}. Furthermore, we provide more analysis and more showcases in Sec.~\ref{appendix:B} and Sec.~\ref{appendix:C}.

\section{More Details}\label{appendix:A}

We provide more details that are not implemented in the main paper. Specifically, we provide the detailed architecture for the model, more training details about \textit{Ola}'s progressively modality alignment, and details of the data preparation procedure. 

\subsection{Model Details}

The visual encoder for \textit{Ola} is based on the SigLIP-400M~\citep{zhai2023siglip} backbone and is further fine-tuned for native-resolution visual inputs. The patch size for the visual encoder is set to 16, the hidden dimension is 1152, and the MLP hidden dimension is 4304. The SigLIP-400M model consists of 27 transformer blocks and 16 attention heads. The audio encoder for \textit{Ola} is built on the whisper-v3-large~\citep{radford2022whisper} and the BEATs~\citep{Chen2022beats} model. The length of the input audio tensor for the audio encoder model is fixed at 480,000; therefore, we chunk the entire audio tensor into pieces and concatenate the audio features. The mel size for the whisper-v3-large model is set to 128, and the hidden dimension for the speech features after the whisper-v3-large model is 1280.

For the connector layer, we utilize a 2-layer MLP for feature projection. We initialize two separate MLPs for visual and audio features, respectively. The input dimension matches that of the visual or audio encoder, and the output dimension matches the LLM dimension. For the \textit{Local-Global Attention Pooling} layer, we use a predictor to calculate the score based on the concatenated features. Therefore, the dimension for the predictor is 2$\times$ to 1$\times$ dimension.

We integrate Qwen-2.5-7B~\citep{qwen2.5} into the \textit{Ola} large language model. The Qwen-2.5-7B model has a hidden LLM dimension of 3,584 and an intermediate size of 18,944. It consists of 28 transformer layers. 
\subsection{Training Details}

As stated in the main paper, the progressive modality alignment procedure is conducted in three stages. The first image-text stage involves adapter pre-training and supervised fine-tuning. The adapter pre-training stage is conducted on 808k image captioning data collected from the LAION datasets. During pre-training, we unfreeze the parameters for the connector while keeping other parameters frozen. We set the training batch size to 256 and the overall learning rate to 1e-3. The supervised fine-tuning stage is conducted on 7.3 million image-text data pairs. During image-text training, input images are maintained at their original aspect ratio, with the maximum image size restricted to 1536×1536. We set the batch size to 128 and the overall learning rate to 2e-5. The stage 1 experiment is conducted on 64 NVIDIA A800 GPUs.

\begin{figure*}[t]
\centering
\includegraphics[width=\textwidth]{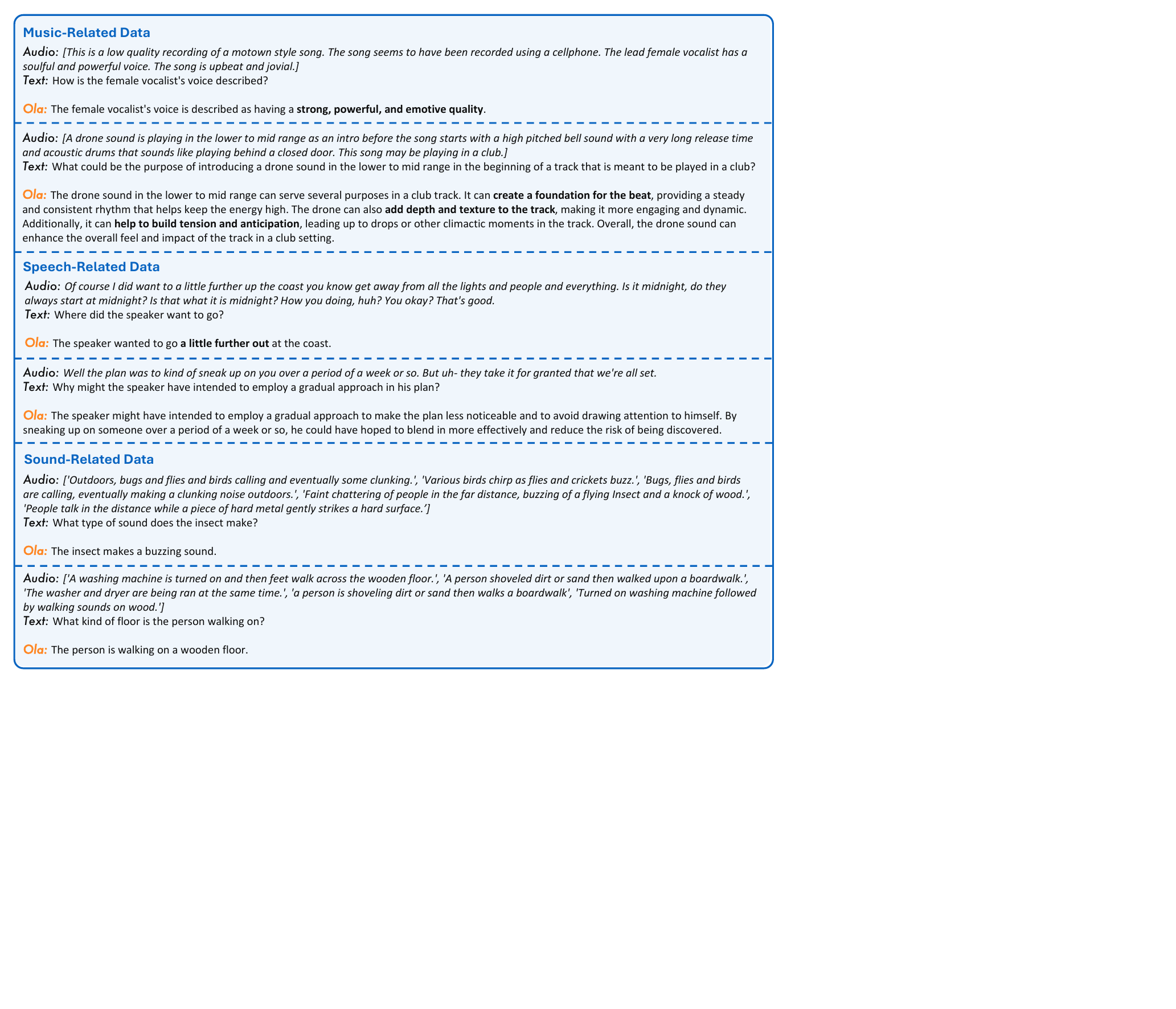}
\caption{\textbf{Showcases on Text and Audio Understanding.}}
\label{fig:supp1}
\end{figure*}

Stage 2 integrates both image and video data for supervised fine-tuning, following most of the training strategies from Stage 1. The total amount of training data is 2.7 million, comprising 800k image-text pairs sampled from Stage 1 and 1.9M video data collected from open-source datasets. We set the training batch size to 256 and the overall learning rate to 2e-5. The model's maximum sequence length is set to 16k. The maximum number of frames is set to 64. The Stage 2 experiment is conducted on 64 NVIDIA A800 GPUs. 

Stage 3 involves joint training in the audio domain. We conduct a projector alignment procedure similar to the image pre-training for initializing the speech adapter. During the speech adapter pre-training, we unfreeze the parameters of the speech adapter while freezing the other parameters. We set the batch size to 256 and the overall learning rate to 1e-3. The pre-training phase is conducted on the 370k LibriTTS~\citep{zen2019libritts} dataset. After pre-training, we integrate audio-video joint alignment by combining image data, video data, and pure audio data. We use 600k image data, 1.1M audio data, and 243k video-audio training data. The training batch size is set to 128, and the overall learning rate is set to 1e-5. We maintain the original audio data for inputs in the audio and video data and append the necessary prompts for instructions. Specifically, for the ASR tasks, we set the ASR prompt as \textit{"Please give the ASR results of the given speech."} For the audio instruction tasks, we set the instruction-following prompt as \textit{"Please directly answer the questions in the user's speech."} We maintain the maximum frame number for the video and set the maximum speech chunk number to 20. The Stage 3 experiment is conducted on 64 NVIDIA A800 GPUs.

\subsection{Data Collection Details}

\begin{figure*}[t]
\centering
\includegraphics[width=\textwidth]{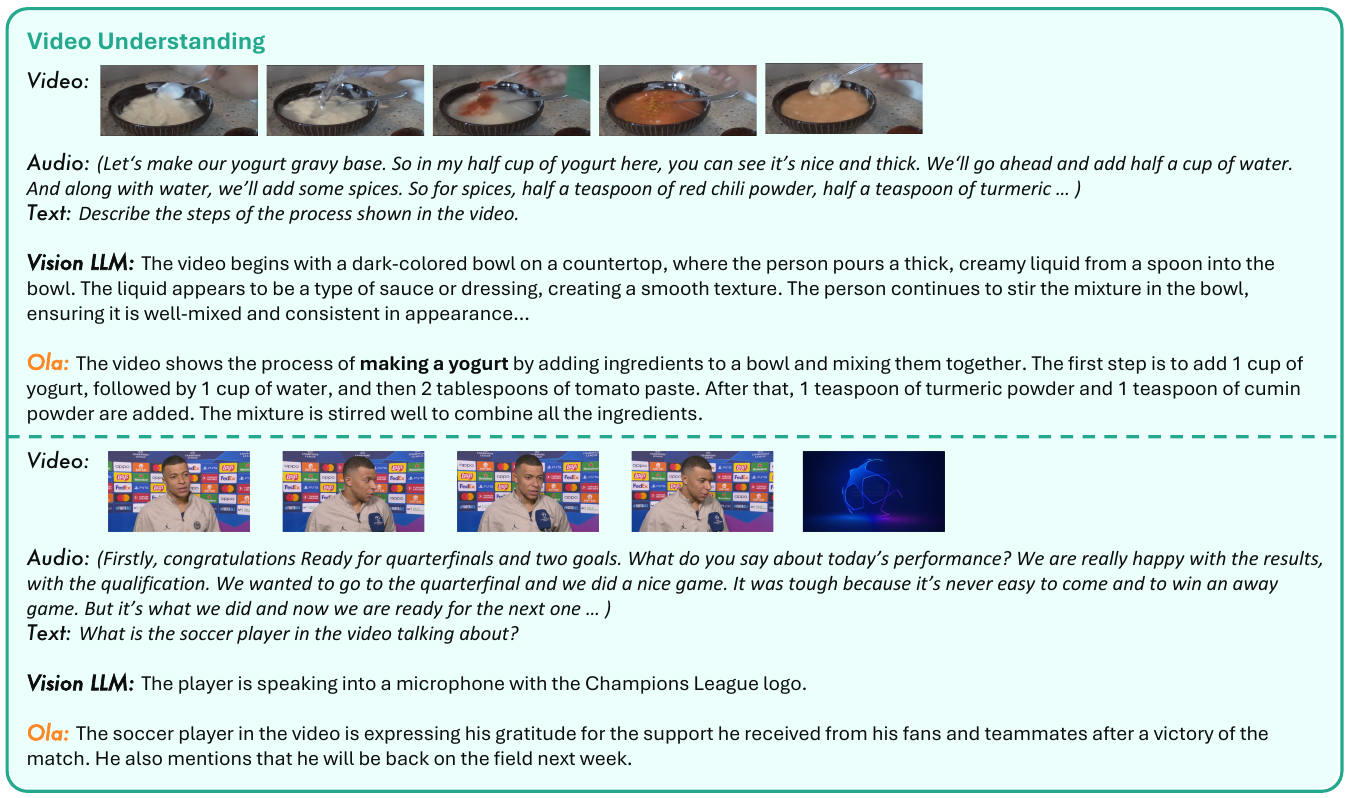}
\caption{\textbf{Showcases on Video Understanding.}}
\label{fig:supp2}
\end{figure*}

In the ViT pre-training phase, we employ Qwen2.5-0.5B as the language interface model to establish cross-modal alignment between visual and textual representations.      The alignment is optimized via cross-entropy loss, which measures the discrepancy between the predicted and ground-truth token distributions over the joint embedding space.      To curate the training data, we aggregate approximately 10M multimodal samples from three primary domains: (1) OCR-intensive datasets (e.g., scene text recognition), (2) visual grounding corpora with region-text pairs, and (3) general vision-language tasks (e.g., image captioning and VQA).   For parameter-efficient adaptation, we integrate a Low-Rank Adaptation (LoRA) module into the LLM component, updating only the low-rank decomposition matrices while freezing the original LLM weights.   In contrast, all parameters of the ViT encoder are fine-tuned end-to-end to preserve the rich visual representations learned during pre-training.  

For the instruction pre-training phase, we curate a multimodal dataset of 20 million image-text pairs from a combination of open-sourced and proprietary in-house datasets, including images from LAION-5B~\citep{schuhmann2021laion}, COYO-700M, Conceptual Captions v3, etc, and knowledge-intensive sources from Wikipedia. To further construct the instruction-quality level data, we use GPT-4o and Gemini-Pro to optimize the question and answer pairs to conduct complexity escalation with multi-hop reasoning requirements and quality unification through style transfer to maintain consistency. We scale up the training batch size to 2048 during the pre-training stage to smooth the gradient and continuously train the model.

Furthermore, we provide details on collecting video-audio relevant data. Our data comes from two sources with high-quality raw videos: LLaVA-Video-178k~\citep{zhang2024llavanextvideo}, which contains 178k raw videos, and FineVideo~\citep{Farré2024FineVideo}, which contains 42k raw videos. For the open-sourced video data from LLaVA-Video-178k, we first use the Whisper~\citep{radford2022whisper} model to generate subtitles. We find that the videos include content in other languages and videos without valid audio, so we design a filtering method for better results. Specifically, we first assess the ratio of English words in the generated subtitles and discard those with a lower ratio, indicating subtitles in other languages. We also discard extremely short subtitles. Then, we use a large language model, Qwen-2.5-72B, to further filter the subtitles. The model is asked to identify meaningless sentences with the following prompt: \textit{"I will give you a subtitle generated from a video. Identify whether the subtitle is complete, fluent, and informative. Answer directly with yes or no and do not add other explanations."} After this procedure, we gathered 41k valid videos. For the videos in FineVideo, as they are already well-processed, we directly use the subtitles for the following steps. We utilize Qwen-2-72B to generate audio-relevant question-answer pairs based on the given videos and subtitles. The prompt for the vision-language model is: \textit{"Please generate at least three questions and answers based on the information in the subtitle. You can refer to the video for additional context. The questions and answers must be highly relevant to the subtitle and video and should not include fabricated content."} We then generate 243k cross-modal video-audio data points from the 81k collected videos. This data is used for stage 3 training for omni-modal alignment.

\section{More Analysis} \label{appendix:B}

\paragrapha{Ablations on Cross-Modal Training Data. }In our implementations, we collect the cross-modal video-audio data for modality alignment from multiple sources including academic datasets and open-ended videos from YouTube. While the data distribution and the processing pipeline vary for the two sources, we conduct ablation analysis on the combination of dual video sources. Results are shown in Tab.~\ref{tab:data} Our baseline model excluded video-audio training, focusing solely on audio-relevant data in Stage 3.  Results indicate that video-audio training minimally affects image benchmarks, suggesting stable image understanding post-text-image training.  For video benchmarks, we observed consistent performance improvements: 59.0\% without video training, rising to 64.2\% with academic data and 65.7\% with open-ended data in VideoMME~\citep{fu2024videomme}. Furthermore, ASR performance on LibriSpeech~\citep{panayotov2015librispeech} improved with video-audio data, likely due to the challenging subtitling tasks in complex environments, enhancing speech recognition capabilities.

% Table generated by Excel2LaTeX from sheet 'Sheet4'
\begin{table}[t]
  \centering
  \caption{\textbf{Analysis on Cross-Modal Training Data.} We analyze our data mixture for the cross-modal video-audio alignment data about sources from academic or open-ended videos. The highlighted row indicates the final accepted strategy. The experiment is conducted on a subsampled training set.} \vspace{5pt}
  \adjustbox{width=\linewidth}{
    \begin{tabular}{cccccc}
    \toprule
    Acadamic & Open-End & ~MMMU~  & VideoMME & LongVideo & LibriSpeech$\downarrow$ \\
    \midrule
    \xmark & \xmark & 48.2  & 59.0 & 56.4  & 4.5 \\
    \cmark & \xmark & \textbf{48.3}  & 64.2  & 56.8  & 4.1  \\
    \rowcolor{Lightorange} \cmark & \cmark & 48.1  &\textbf{ 65.7}  & \textbf{57.4}  & \textbf{4.0}  \\
    \bottomrule
    \end{tabular}%
    }
  \label{tab:data}%
\end{table}%

\section{More Showcases} \label{appendix:C}

\subsection{Text and Audio Understanding}

In this subsection, we provide more practical text-audio understanding samples for visualizations. The inputs of the text-audio understanding are a mixture of audio and text instructions, which can strongly test the cross-modal capability for \textit{Ola} model. Results are shown in Fig.~\ref{fig:supp1}, we provide results on music-related, speech-related, and sound-related audio inputs and \textit{Ola} excels at all the circumstances with a strong performance on mixed audio and text understanding. 

\subsection{Video Understanding}

In this subsection, we provide more results on video understanding tasks and provide comparisons with state-of-the-art vision LLM. Results are shown in Fig.~\ref{fig:supp2}. With the capability to recognize video, audio, and text jointly, \textit{Ola} can gather more information from the video. 

\section{Limitations}
In this section, we discuss the limitations of our work. Ola focuses solely on text generation for omni-model understanding and does not explore audio generation. This approach limits the model's ability to fully leverage multimodal inputs, potentially restricting its applicability in scenarios where audio generation is crucial. Future work could address this by incorporating audio generation capabilities to enhance the model's comprehensive understanding across different modalities.

{
    \small
    \bibliographystyle{ieeenat_fullname}
    \bibliography{main}
}

\end{document}